\documentclass[conference]{IEEEtran}
\IEEEoverridecommandlockouts
\usepackage{amsmath,amssymb,amsfonts}
\usepackage{algorithmic}
\usepackage{graphicx}
\usepackage{textcomp}
\usepackage{xcolor}
\usepackage{forest}
\usepackage{booktabs}
\usepackage{array}
\usepackage{url}
\usepackage{tikz}
\usepackage[square,numbers]{natbib}
\usepackage{caption}
\usepackage{verbatim}
\definecolor{linecol}{RGB}{0,0,0} % Black

\newcommand{\Int}{{\texttt{InceptionTime}}{} }
\newcommand{ \rocket }{{\texttt{ROCKET}}{} }

\def\BibTeX{{\rm B\kern-.05em{\sc i\kern-.025em b}\kern-.08em
    T\kern-.1667em\lower.7ex\hbox{E}\kern-.125emX}}

\colorlet{level1}{blue!50}
\colorlet{level2}{red!50}
\colorlet{level3}{green!50}
\colorlet{linecol}{black!75}

\definecolor{basicTech}{RGB}{255, 182, 193}  % Light Pink for basic techniques
\definecolor{genTech}{RGB}{188, 143, 143}    % Rosy Brown for generative techniques, a soft brown
\definecolor{presTech}{RGB}{216, 191, 216}   % Thistle for preserving techniques, a light purple

\definecolor{linecol}{RGB}{0,0,0}            % Black for lines

\tikzset{
  my rounded corners/.style={rounded corners=2pt}
}
\forestset{
  my tree/.style={
    for tree={
      line width=1pt,
      draw=linecol,
      edge={color=linecol, >={}, ->},
      align=center,
      font=\sffamily\small,
      tikz={my rounded corners},
      if level=0{
        l sep+=5pt,
        calign=child,
        calign child=2,
        for descendants={
          align=center
        },
      }{},
      if level=2{
        for descendants={
          child anchor=west,
          parent anchor=west,
          align=center,
          anchor=west,
          edge path={
            \noexpand\path[\forestoption{edge}]
            (!u.parent anchor) -- ++(0,-10pt) -| (.child anchor)\forestoption{edge label};
          },
        },
      }{},
    }
  }
}

\begin{document}

\title{Data Augmentation for Multivariate Time Series Classification: An Experimental Study
}

\renewcommand{\footnoterule}{%
    \hrule width \linewidth height 0.5pt
    \vspace{4pt}
}

\author{%
  Romain Ilbert\thanks{Corresponding author: \texttt{romain.ilbert@hotmail.fr}.}$^{*1,2}$ \quad Thai V. Hoang$^3$ \quad Zonghua Zhang$^4$\\
  $^1$Huawei Noah’s Ark Lab, Paris, France \quad $^2$LIPADE, Paris Descartes University, Paris, France \\ $^3$ TH Consulting, Paris, France $^4$ CRSC R\&D Institute Group Co. Ltd, Beijing, China
}

\newcommand{\tp}[1]{{\color{red} {\bf ??? #1 ???}}\normalcolor}
\newcommand{\ro}[1]{{\color{orange} {\bf // #1 //}}\normalcolor}

\maketitle

\begin{abstract}
Our study investigates the impact of data augmentation on the performance of multivariate time series models, focusing on datasets from the UCR archive. 
Despite the limited size of these datasets, we achieved classification accuracy improvements in 10 out of 13 datasets using the \rocket and InceptionTime models. 
This highlights the essential role of sufficient data in training effective models, paralleling the advancements seen in computer vision.
%We extend the realm of data augmentation to multivariate time series classification, introducing innovative methods specifically designed for time series data. 
%\ro{Our work delves into adapting and applying existing methods in innovative ways to the domain of multivariate time series classification.}
Our work delves into adapting and applying existing methods in innovative ways to the domain of multivariate time series classification.
Our comprehensive exploration of these techniques sets a new standard for addressing data scarcity in time series analysis, emphasizing that diverse augmentation strategies are crucial for unlocking the potential of both traditional and deep learning models.
Moreover, by meticulously analyzing and applying a variety of augmentation techniques, we demonstrate that strategic data enrichment can enhance model accuracy. 
This not only establishes a benchmark for future research in time series analysis but also underscores the importance of adopting varied augmentation approaches to improve model performance in the face of limited data availability.
\end{abstract}


\section{Introduction}
The progression of machine learning, especially in time series classification, has been markedly accelerated by the advent of deep learning techniques.
These models, however, exhibit an inherent dependency on large and diverse training datasets to achieve optimal performance, a requirement often challenging to meet in practice~\cite{modern_nn}. 
The scarcity and imbalance of classes in these datasets, poses a critical bottleneck, affecting not only model accuracy but also their ability to generalize across diverse scenarios~\cite{NLP, survey_image}.

In the fields of computer vision and natural language processing (NLP), data augmentation has emerged as a fundamental technique, effectively addressing the limitations posed by insufficient data~\cite{dl_image_classif, data_aug_nlp}.
By artificially enhancing dataset size and diversity, data augmentation techniques have proven to significantly mitigate overfitting, thereby improving model robustness and performance~\cite{survey_image}. 
This success has sparked interest in applying similar strategies within the domain of time series classification, where the challenges of data scarcity and class imbalance are equally prevalent~\cite{ts_data_aug_dl_survey, survey_ts_classif, fault_pred_aug, anomaly_aug}.

Class imbalance, in particular, is a pervasive issue that skews the learning process, often resulting in models that are biased towards the majority class \cite{imbalanced_classes}. 
Addressing this imbalance is critical, especially in multivariate time series datasets where the complexity and variability of data exacerbate the problem. 
Our study concentrates on the UCR archive, which has recently been enriched with a broad array of multivariate time series datasets, offering an ideal environment for investigating the effectiveness of data augmentation within this domain~\cite{ucr, uea}.

Our work incorporates both traditional and deep learning approaches, namely, the \rocket and \Int models. 
These models represent the state-of-the-art in time series classification, offering a unique blend of speed, accuracy, and adaptability across a wide range of time series data~\cite{rocket, inception_time, bake_off_2017, bake_off_2021}. 
The inclusion of these models allows us to comprehensively evaluate the impact of data augmentation on both traditional and deep learning approaches, ensuring our findings are broadly applicable and relevant to current classification challenges~\cite{ts_chief, hive_cote_2}.

Central to our investigation is a detailed exploration of data augmentation techniques tailored specifically for time series data. 
Among these, the Synthetic Minority Over-sampling Technique (SMOTE), and a noise injection, stand out for their ability to generate synthetic data that closely mimics the original datasets, thus addressing the dual challenges of data scarcity and class imbalance~\cite{SMOTE}. 
While these methods can be applied to both univariate and multivariate time series, we also explore the potential of  Time Generative Adversarial Networks (TimeGANs) for their ability to capture complex inter-variable dependencies~\cite{TimeGAN}. 
This makes them a promising candidate for multivariate time series analysis, and we evaluate them in our work.
Our methodology encompasses a diverse range of augmentation strategies, each carefully selected to enhance the representativeness and quality of the training data, thereby enabling models to achieve superior generalization and performance~\cite{ts_data_aug_dl_survey}.

In this study, we rigorously explored a wide range of data augmentation techniques, meticulously selected from the diverse branches of our newly developed taxonomy. By combining this approach with an in-depth analysis of two leading time series classification methodologies (i.e., \rocket and \Int), not only do we demonstrate improvements in model accuracy, %across 13 imbalanced multivariate datasets from the UCR/UEA archive, 
but also shed light on the complex interplay between data characteristics, augmentation strategies, and model performance. 
Our results indicate that accuracy enhancements are not the result of any single augmentation technique, but rather emerge from a combination of methods, highlighting the lack of a one-size-fits-all solution in applying specific augmentation strategies.

%Beyond demonstrating the immediate benefits of data augmentation on model accuracy, our research delves into the strategic selection and combination of augmentation techniques, drawing parallels with practices in computer vision, where a mix of methods has significantly improved classification results. 
The results of our comprehensive investigation advocate for an informed use of data augmentation in time series classification. This work contributes to the academic and practical discourse on overcoming challenges like data scarcity and class imbalance, and also paves the way for future advancements in this area.

Our contributions can be summarized as follows.

\begin{itemize}
    \item We expand the understanding of time series classification by evaluating both \Int and the state-of-the-art \rocket models, highlighting the importance of considering both deep learning and non-deep learning approaches.
    
    \item We conduct an exhaustive review of data augmentation techniques and introducing a new taxonomy (Figure~\ref{fig:taxonomy}) that categorizes these methods into distinct branches, enriching the framework for their application and evaluation.
    
    \item We utilize a variety of augmentation techniques from different branches of our taxonomy, including those necessitating external training like TimeGANs, marking a first in the context of time series data augmentation.
    
    \item We demonstrate accuracy improvements through empirical evaluation on the 13 multivariate, imbalanced datasets from the UCR/UEA archive.
    
    \item Our detailed analysis reveals that a broad spectrum of augmentation techniques can enhance model accuracy, underscoring the variability in their effectiveness across datasets and suggesting the potential for optimization through strategic combination.
\end{itemize}

Our study establishes a foundation for future research aimed at refining the application of data augmentation in time series classification. 
Inspired by successful strategies in computer vision, we believe that exploring the synergistic use of varied augmentation techniques can lead to further performance improvements.

% The rest of this paper is organized as follows: Section~\ref{sec:taxonomy} introduces a comprehensive taxonomy of data augmentation techniques, which are further elaborated in Section~\ref{sec:overview}. 
% Our experimental evaluation is detailed in Section~\ref{sec:exp}, followed by conclusions in Section~\ref{sec:conclusion}.

\section{A Taxonomy of Time Series Augmentation Techniques}
\label{sec:taxonomy}

\begin{figure*}[!t]
\centering
\scalebox{0.72}{
\begin{forest}
  for tree={
    line width=1pt,
    draw=linecol,
    edge={color=linecol, >={}, ->},
    align=center,
    font=\sffamily\small,
     % Applying rounded corners to all nodes
    % Define spacing and alignment adjustments
    if level=0{
      l sep+=5pt,
      calign=child,
      calign child=2,
      for descendants={
        align=center
      },
    }{},
    % Custom adjustments for level 2 nodes
    if level=2{
      for descendants={
        child anchor=west,
        parent anchor=west,
        align=center,
        anchor=west,
        edge path={
          \noexpand\path[\forestoption{edge}]
          (!u.parent anchor) -- ++(0,-10pt) -| (.child anchor)\forestoption{edge label};
        },
      },
    }{},
  },
  [Time Series Data Augmentation Techniques, fill=gray!30
    [Basic Techniques, for tree={fill=basicTech}
      [Time\\ Domain, for tree={fill=basicTech!90}
        [Slicing\\ \cite{Data,survey_ts_classif, ts_data_aug_dl_survey}
        [Permutation\\ \cite{survey_ts_classif, ts_data_aug_dl_survey}
        [Warping\\ \cite{Data, Guide, survey_ts_classif, ts_data_aug_dl_survey, time_warp}
        [Masking\\ \cite{Cutout, SpecAugment}
        [Injecting Noise\\ \cite{noise_injection, Tikhonov, lstm, survey_ts_classif, ts_data_aug_dl_survey}
        [Rotation\\ \cite{survey_ts_classif, ts_data_aug_dl_survey, tdnn_lstm}
        [Scaling\\ \cite{survey_ts_classif, ts_data_aug_dl_survey}]
      ]]]]]]]
      [Frequency\\ Domain, for tree={fill=basicTech!80}
        [Fourier\\ Transform\\ \cite{survey_ts_classif, fourier_average}
        [Frequency\\ Warping\\ \cite{survey_ts_classif, VTLP}
        [Frequency\\ Masking\\ \cite{survey_ts_classif, SpecAugment}
        [Mixing\\ \cite{survey_ts_classif, AER, SFM}]
      ]]]]
      [Oversampling\\ Techniques, for tree={fill=basicTech!70}
        [Interpolation\\ \cite{SMOTE, ANSMOT, SMOTEFUNA, borderline}
        [Density\\ \cite{ADASYN, SWIM}]
      ]]
      [Decomposition\\ Techniques, for tree={fill=basicTech!60}
        [STL\\ \cite{survey_ts_classif, RobustSTL}
        [EMD\\ \cite{survey_ts_classif, EMD, emd_aug}
        [RobusTAD\\ \cite{survey_ts_classif, Robusttad}
        [ICA\\ \cite{ICA_aug, ICA}]]]]
      ]
    ]
    [Generative Techniques, for tree={fill=genTech}
      [Statistical\\ Models, for tree={fill=genTech!90}
        [Posterior\\ Sampling\\ \cite{survey_ts_classif, posterior}
        [Gaussian\\ Trees\\ \cite{survey_ts_classif, Mixture}
        [LGT\\ \cite{survey_ts_classif, LGT}
        [GRATIS\\ \cite{survey_ts_classif, GRATIS}]]]]
      ]
      [Neural\\ Networks, for tree={fill=genTech!80}
        [Autoencoders\\ \cite{survey_ts_classif, latent_efficient, latent_space, mixup} \\ \cite{ modals, LSTMAE, DTWSSE, data_aug_ts_gen}
        [GANs\\ \cite{survey_ts_classif, GAN, TimeGAN, WaveGAN, wgan_first} \\ \cite{WGAN-VAEs, TCGAN, bio_GANS, emotionalGAN}]]
      ]
      [Probabilistic\\ Models, for tree={fill=genTech!70}
        [Autoregressive\\ Models\\ \cite{Wavenet, deep_ar}
        [Diffusion\\ Models\\ \cite{diffusion}
        [Normalizing\\ Flows\\ \cite{normalizing_flows, dynamic_norm_flows, VAEs_flow}]]]
      ]
    ]
    [Preserving Techniques, for tree={fill=presTech}
      [Label\\ Preserving, for tree={fill=presTech!90}
        [Range\\ Techniques\\ \cite{noise_standard, parkinson, class_preserv}]
      ]
      [Structure\\ Preserving, for tree={fill=presTech!80}
        [SPO\\ \cite{SPO}
        [INOS\\ \cite{INOS}
        [MDO\\ \cite{MDO}
        [OHIT\\ \cite{oversampling_imbalanced}]]]]
      ]
    ]
  ]
\end{forest}
}
\caption{Comprehensive taxonomy of data augmentation techniques for time series analysis, integrating a wide array of methodologies from basic transformations to advanced generative models, including a branch on Preserving Techniques.}
\label{fig:taxonomy}
\end{figure*}
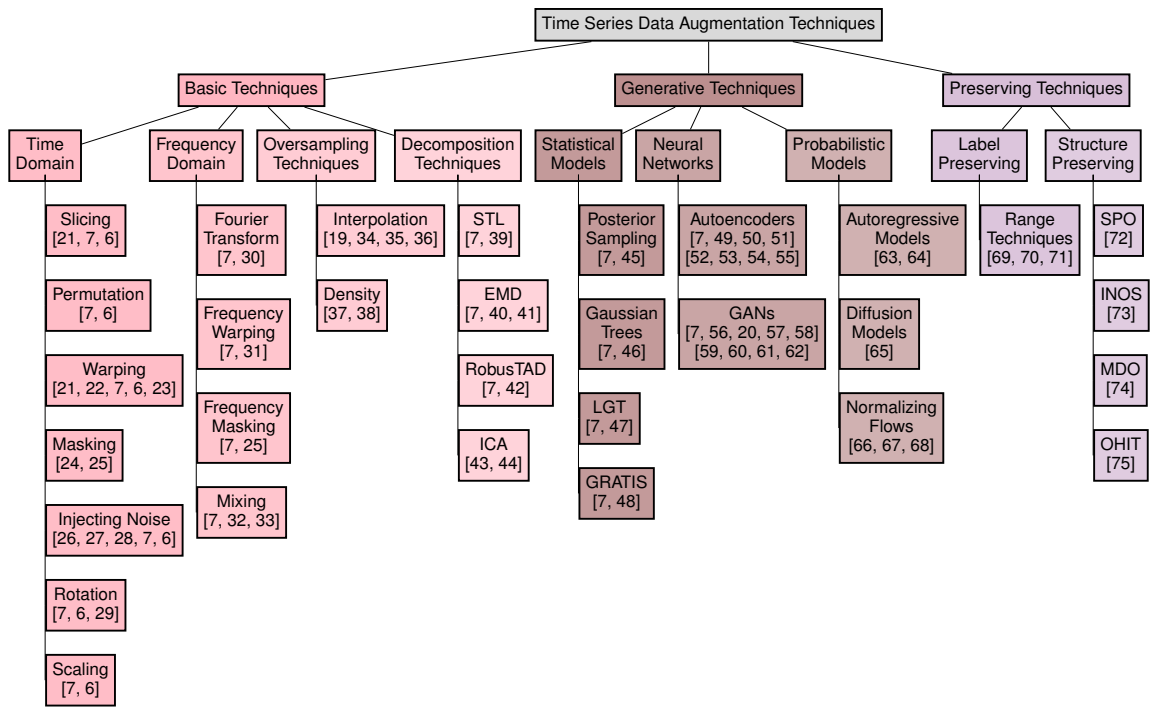
Figure~\ref{fig:taxonomy} presents a comprehensive taxonomy of data augmentation techniques, which we discuss in Section~\ref{sec:overview}.

\begin{figure*}[tb]
\centering
\scalebox{0.82}{
\begin{minipage}[b]{0.31\linewidth}
\centering
\includegraphics[width=0.83\linewidth]{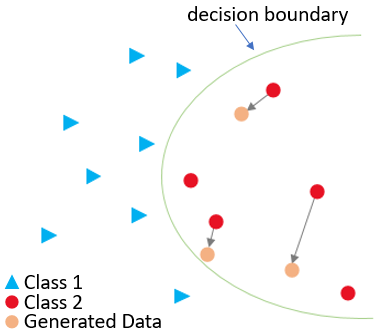}
\vspace{.3cm}
\caption{Basic Techniques, like noise~injection}
\label{fig:basic}
\end{minipage}
\hspace{0.3cm}
\begin{minipage}[b]{0.31\linewidth}
\centering
\includegraphics[width=0.92\linewidth]{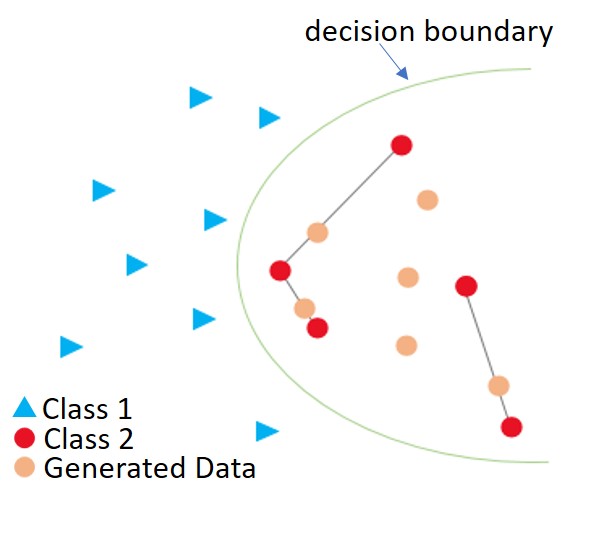}
\caption{Oversampling Techniques, like SMOTE}
\label{fig:interp}
\end{minipage}
\hspace{0.3cm}
\begin{minipage}[b]{0.31\linewidth}
\centering
\includegraphics[width=0.98\linewidth]{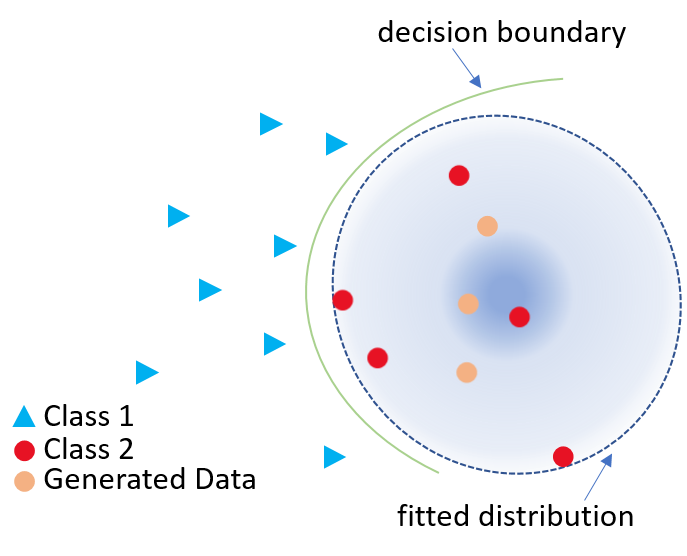}
\vspace{-0.6cm}
\caption{Generative Techniques, like timeGANs}
\label{fig:gen}
\end{minipage}}

\medskip
\centering
\scalebox{0.82}{
\centering
\begin{minipage}[b]{0.35\linewidth}
\centering
\includegraphics[width=\linewidth]{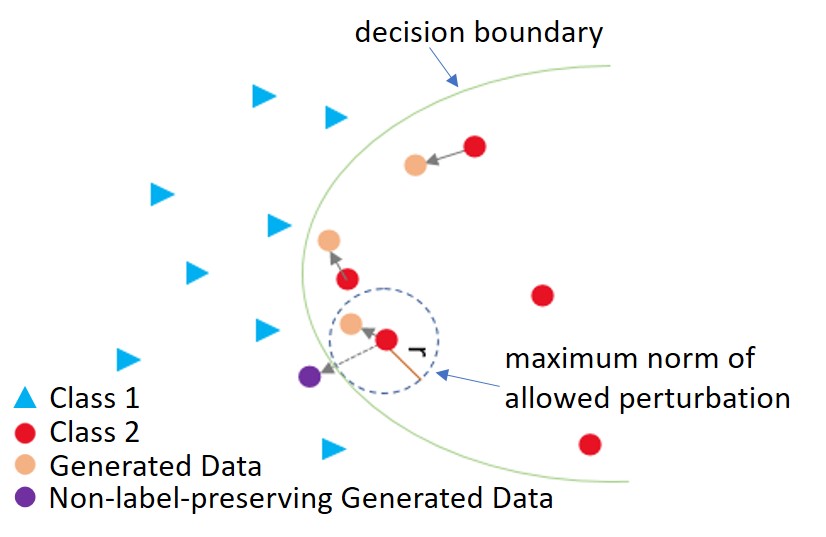}
\caption{Label-Preserving Techniques, like range techniques}
\label{fig:label-preserv}
\end{minipage}
\hspace{2.5cm}
\begin{minipage}[b]{0.31\linewidth}
\centering
\includegraphics[width=0.89\linewidth]{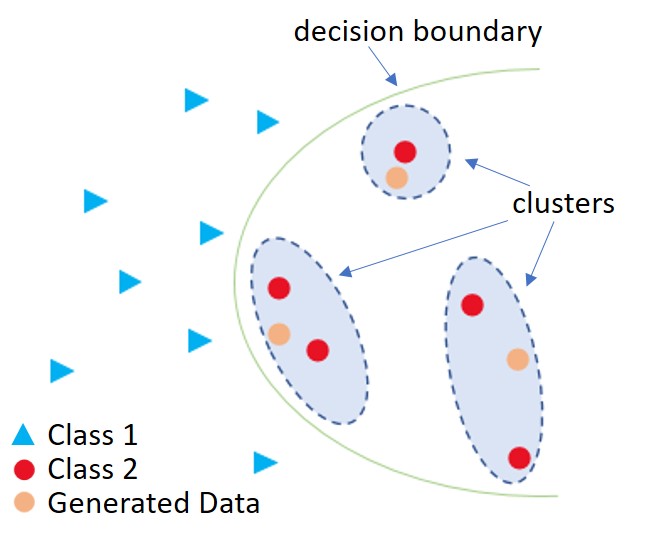}
\vspace{0.05cm}
\caption{Structure-Preserving Techniques like OHIT}
\label{fig:struct-preserv}
\end{minipage}
}
\end{figure*}

The initial category encompasses \emph{basic} techniques, which include methods such as slicing, cropping, or noise injection, applicable within both the \emph{time} and \emph{frequency} domains (see Figure~\ref{fig:basic}). 
This group also includes techniques based on \emph{oversampling} and \emph{decomposition}.

Subsequently, the \emph{generative} class of techniques is comprised of methods subdivided into \emph{statistical}, \emph{neural network}, and \emph{probabilistic} approaches. 
These strategies aim to emulate the authentic probability distribution of the time series data to generate new instances.

The final class, \emph{preserving}, maintains the original classes found within the dataset. 
It is further segmented into two sub-categories: \emph{label-preserving}, which fine-tunes common techniques such as noise injection to preserve accuracy, and \emph{structure-preserving}, which focuses on maintaining the spatial structure and inter-point dependencies within the data.

For an overarching view of the taxonomy and to discern the differences between the approaches under the various branches, the reader is directed to Figures~\ref{fig:basic}-\ref{fig:struct-preserv}.

In Figure~\ref{fig:basic}, we demonstrate the technique of \emph{noise injection}~\cite{noise_injection, Tikhonov, lstm, survey_ts_classif}, a foundational data augmentation method that modifies data points in the time domain. 
Another representative of basic augmentation methods, depicted in Figure~\ref{fig:interp}, is the SMOTE algorithm~\cite{SMOTE}, which fabricates new instances by creating convex combinations of pre-existing examples within the dataset.
Turning to generative strategies, Figure~\ref{fig:gen} portrays the TimeGANs technique~\cite{TimeGAN}, a generative neural-network-based method. 
The overarching aim of generative techniques is to construct a model that approximates the minority class distribution, which can subsequently be used to produce novel time series data. The primary distinction among these generative approaches is their respective methodologies for approximating the distribution of the minority class.
An exemplification of a label-preserving approach is depicted in Figure~\ref{fig:label-preserv}, showcasing a range method where the paramount objective is to ensure that the newly generated data points do not transgress the decision boundary. 
This is a potential issue with elementary noise-injection techniques, such as the one shown in Figure~\ref{fig:basic}, where a straightforward application of noise can inadvertently shift data points across the decision boundary. 
Range techniques meticulously modulate the extent of noise application to guarantee adherence to the decision boundary.
Lastly, Figure~\ref{fig:struct-preserv} illustrates an example of a structure-preserving technique, notably OHIT~\cite{oversampling_imbalanced}. 
This approach creates clusters and computes their covariance matrices, which are then used to generate new examples that are likely to fall within the boundaries of these clusters.

Our taxonomy sets itself apart from other taxonomies~\cite{survey_ts_classif, ts_data_aug_dl_survey} by incorporating the \emph{preserving} class of techniques, which try to address the following challenges.
First, when performing data augmentation by adding noise, how can we determine the optimal amount of noise to augment a series intelligently?
Second, if our original time series are interdependent (e.g., correlated), how can we generate new series that retain these dependencies? 
Additionally, we introduce in the taxonomy the probabilistic models (under the \emph{generative} class of techniques), which describe time series as transformations of underlying Markov processes that are easier to model.
Finally, we include a new neural-network model that require external training, the TimeGANs.

\section{Overview of Time Series Augmentation Techniques}
\label{sec:overview}

A Multivariate Time Series (MTS), denoted as \(x=(x_1,...,x_t,...,x_T)\), is composed of \(T\) sequentially ordered elements, where each element \(x_t\) resides in an \(M\)-dimensional space, i.e., \(x_t \in \mathbb{R}^M\). 
We will use the term \emph{data point} to refer to an individual observation within the series, which has \(M\) dimensions, or to the entire series, which encompasses \(T\) dimensions, contingent on the specific analytical context.
%\ro{It's worth noting that the approaches and techniques discussed here, while focused on multivariate time series, could potentially be adapted and applied to univariate time series as well.}
It's worth noting that the approaches and techniques discussed here, while focused on multivariate time series, could potentially be adapted and applied to univariate time series as well.

\subsection{Basic Techniques}
\label{subsec:basic_techniques}

\subsubsection{Time Domain}
\label{subsubsec:time-domain}
Time domain augmentation involves modifying time or magnitude. Techniques include noise injection for regularization~\cite{noise_injection, Tikhonov, lstm, survey_ts_classif, ts_data_aug_dl_survey}, scaling for magnitude adjustment~\cite{survey_ts_classif, ts_data_aug_dl_survey}, rotation affecting temporal dependencies~\cite{survey_ts_classif, ts_data_aug_dl_survey, tdnn_lstm}, slicing for segment extraction~\cite{Data, survey_ts_classif, ts_data_aug_dl_survey}, permutation of series intervals~\cite{survey_ts_classif, ts_data_aug_dl_survey}, and regularization methods like masking, cropping, dropout, and pooling~\cite{Cutout, SpecAugment}. 
Window Warping and guided warping use temporal distortions and Dynamic Time Warping for novel series generation~\cite{Data, Guide, survey_ts_classif, ts_data_aug_dl_survey, time_warp, DTW, ShapeDTW, dba}.

\subsubsection{Frequency Domain}
\label{subsubsec:frequency-domain}
Frequency domain augmentation applies amplitude and phase perturbations~\cite{Robusttad}, with STFT for spectrograms~\cite{fourier_average}. 
EMDA perturbs frequency characteristics for Acoustic Event Detection~\cite{AER}. 
VTLP and SFM distort speech spectra, or convert speech data~\cite{VTLP, SFM}. % for voice conversion~\cite{VTLP, SFM}.

\subsubsection{Oversampling Techniques}
\label{subsubsec:Oversampling}
Oversampling treats time series as spatial points for augmentation. 
Interpolation mixes a series with its nearest neighbor~\cite{SMOTE}. 
SMOTE and its variants—ANSMOT and SMOTEFUNA—along with ADASYN and SWIM, address minority class enhancement through density-based synthetic sample generation~\cite{ANSMOT, SMOTEFUNA, ADASYN, SWIM, borderline}.

\subsubsection{Decomposition-Based Techniques}
\label{subsubsec:decomposition}
Time series can be decomposed into trend, seasonality, and residual components for targeted augmentation. 
Techniques include RobustTAD for anomaly detection~\cite{Robusttad}, RobustSTL for seasonality handling~\cite{RobustSTL}, EMD for sensor data noise reduction~\cite{EMD, emd_aug}, and ICA with D-FANN for series gap filling~\cite{ICA_aug, ICA}.

\noindent Combining techniques like permutation, rotation, time warping~\cite{parkinson}, and SpecAugment's spectrogram operations—time warping, frequency, and time masking—can optimize augmentation~\cite{SpecAugment}.

%%--------------------------------------------------------------------------------------------------------------

%-----------------------------------------------------------------------------------------------------------
\subsection{Generative Techniques Overview}

Time series can be sampled directly from a posterior distribution, as depicted in Figure~\ref{fig:gen}. 
This section delves into two main types of generative models: statistical and neural network-based models.

\subsubsection{Statistical Generative Models}
Recent years have seen significant interest and advancements in generative models. 
Tanner and Wong~\cite{posterior} suggested approximating the true posterior distribution for generating new variables. 
Research by \cite{Mixture} leverages the strong correlation between close time points. 
Bellman~\cite{Bellman} utilized sparse graphical models to capture statistical dependencies over time. 
Smyl and Kuber~\cite{LGT} showcased the effectiveness of Local and Global Trend (LGT) data augmentation, particularly when combined with LSTMs. 
GRATIS~\cite{GRATIS} investigated time series characteristics and time dependency to efficiently produce new series. 
Vinod et al.~\cite{Bootstrap} implemented a maximum entropy bootstrap method for generating instances closely related to the originals. 
Moreover, \cite{NNAutoML} proved that combining data augmentation with neural architecture exploration yields promising outcomes.

\subsubsection{Neural Networks Based Generative Models}
\label{subsubsec:nn-generative-models}

This section reviews neural network architectures for augmenting time series data. 
Auto-encoders (AE) leverage a latent space for efficient transformations like interpolation~\cite{latent_space, mixup}, outperforming direct raw input use~\cite{latent_efficient}. MODALS~\cite{modals} automates augmentation, while LSTM auto-encoders (LSTM-AE)~\cite{LSTMAE} enhance spatial-temporal data. Variational auto-encoders (VAE) and conditional VAEs, as shown by Kirchbuchner et al.~\cite{data_aug_ts_gen}, effectively reduce target data variance. 
Combining LSTM-based VAE samples with interpolation~\cite{latent_space} augments time series, with Qingsong Wen et al.~\cite{ts_data_aug_dl_survey} evaluating DeepAR and transformer-based techniques. 
DTW-based SMOTE with Siamese Encoders (DTWSSE)~\cite{DTWSSE} and generative adversarial networks (GANs)~\cite{GAN} also contribute to augmentation, including MLP, RNN~\cite{bio_GANS, traffic}, 1D CNN~\cite{TCGAN, emotionalGAN}, and 2D CNN~\cite{WaveGAN} variants. 
The WGAN discriminator replaces the VAE decoder for enhanced performance in~\cite{WGAN}, with selective WGAN (sWGAN) and VAE (sVAE) outperforming conditional WGANs (cWGAN)~\cite{WGAN-VAEs}. DOPING~\cite{doping} utilizes adversarial autoencoders (AAE)~\cite{AAE} for oversampling, while TimeGANs~\cite{TimeGAN} aim to preserve temporal series dynamics.

\subsubsection{Probabilistic Models}
\label{subsubsec:proba-models}

Generative models also augment time series data. Wavenet~\cite{Wavenet}, a deep probabilistic autoregressive NN, generates raw audio by factorizing the probability distribution as:
\begin{equation}
   \mathbb{P}(x) = \prod\limits_{t=1}^T \mathbb{P}(x_t | x_1,...,x_{t-1})	
\end{equation}

GluonTS~\cite{ts_modeling} offers transformer and Wavenet implementations. 
DeepAR~\cite{deep_ar} trains an autoregressive RNN for probabilistic forecasting. 
Normalizing flows~\cite{normalizing_flows}, introduced by Brubaker et al.~\cite{dynamic_norm_flows}, map simple distributions to complex ones via invertible, differentiable mappings. 
They used a VAE to initialize a base distribution for training a normalizing flow~\cite{VAEs_flow}. 
Diffusion Models, through a Markov chain of diffusion steps, gradually introduce then remove noise, learning to recreate the original data from the noise, focusing on the conditional backward probability.

\begin{equation}
   \mathbb{P}_\theta(x) = \mathbb{P}(x_T) \prod\limits_{t=1}^{T-1}\mathbb{P}_\theta(x_{t-1} |x_t)	
\end{equation}
where $\mathbb{P}_\theta(x_{t-1} |x_t) \sim \mathcal{N}(\mu_\theta(x_t,t),\Sigma_\theta(x_t,t))$.

%-----------------------------------------------------------------------------------------------------------
\subsection{Structure- and Label-Preserving Techniques}

In the case of sensor signals, collecting a large amount of data samples under various operating conditions, or from different environments, is a complex task.
Data augmentation is a solution to address this problem: different transformations are applied on the data, in order to create new data points. 

However, the labels of these new points may often times be sensitive to even small fluctuations of the points' values. 
Evidently, we do not want to produce new data points that, even though in the neighborhood of an existing class, lie on the other side of the class decision boundary (refer to Figure~\ref{fig:label-preserv}). 
Moreover, sensor data make it hard for a human analyst to recognize differences in the labels between raw and augmented signals (unlike image classification for example, where visual inspection is an effective solution).
To resolve this issue, we need to make sure that the generated data have the right label, as well as follow the same data characteristics (as the rest of the points in the same neighborhood in the data space).

\subsubsection{Label-preserving}
\label{subsubsec:label-preserving}

Augmentation techniques must preserve labels to avoid misclassification, such as false positives from noise in Parkinson's disease analysis, where noise could mimic dyskinesia symptoms, degrading performance~\cite{parkinson}. 
Cropping risks losing critical information like shapelets, detrimental in small datasets~\cite{shapelet, parkinson}. Classification can be misled by scaling in datasets where intensity distinguishes labels~\cite{parkinson}. 
It's crucial to understand class-specific regions to determine safe perturbation amplitudes (see Figure~\ref{fig:label-preserv}), enhancing test accuracy by 5\% without model adjustments~\cite{class_preserv}.

\subsubsection{Structure-preserving}
\label{subsubsec:structure-preserving}

Research has explored SNN-based density clustering for high-dimensional data, addressing MDO's shortcomings in estimating the true covariance matrix~\cite{SNN, MDO}. 
OHIT addresses high-dimensional, imbalanced time-series classification by using similarity-based clustering to reflect minority class modality, generating new samples to preserve mode covariance structures (Figure~\ref{fig:struct-preserv})~\cite{oversampling_imbalanced}. 
INOS introduces structure-preserving oversampling for imbalanced time series by first generating samples via interpolation, then creating additional synthetic samples based on a regularized minority class covariance matrix, enhancing SPO~\cite{INOS, SPO}.

\section{Experimental Evaluation}
\label{sec:exp}

%\subsection{Setup}
\label{subec:setup}

In this research, Python 3.7.3 served as the primary programming language. 
The implementation of SMOTE relied on the imbalanced-learn library (version 0.8.0). 
Modifications to TimeGANs were carried out using the ydata-synthetic package (version 0.7.1) alongside Tensorflow (version 2.4.4). 
Noise injection was facilitated through numpy (version 1.19.2). 
Classification tasks utilized sktime (version 0.13.0), sklearn (version 1.0.1), fastai (version 2.7.7), and tsai (version 0.3.1), all of which underwent slight modifications for this study. 
Computations were performed on the Jean Zay supercomputer, equipped with NVIDIA V100 GPUs boasting 16 GB of RAM.

We make all code used in this paper available online: \url{https://helios2.mi.parisdescartes.fr/~themisp/tsda/} .

%-----------------------------------------------------------------------------------------------------------
\subsection{Baseline Algorithms}
\label{subsec:Baseline_Algorithms}

In time series classification, dataset imbalances between minority (positive) and majority (negative) samples are common, necessitating dataset augmentation to improve minority representation. 
This field focuses on categorizing data sequences by temporal patterns, crucial for binary classification (normal vs. abnormal sequences) and multi-category scenarios.

Advancements in model performance are notable. A study~\cite{bake_off_2017} highlights top time series classification techniques using intervals, shapelets, or word dictionaries, while another review~\cite{dl_classif_review} examines deep learning approaches in this area.

It appears from the above two studies that the best classification models are COTE~\cite{cote} for non deep learning models, and models with residual connections for deep learning ones~\cite{resnet}.
The COTE algorithm was later improved in HIVE-COTE~\cite{hive_cote_2016, hive_cote_2018} and HIVE-COTE 2.0 (HC2)~\cite{hive_cote_2}, while Resnet became a basis for \Int~\cite{inception_time}.
\cite{ts_chief} proposed a novel time series classification algorithm, TS-CHIEF, which rivals HIVE-COTE in accuracy but requires only a fraction of the runtime. 
Then, a new family appeared: \rocket~\cite{rocket}, which has the advantage of being very fast, compared to the HIVE-COTE algorithm.
\cite{bake_off_2021} gives an overview of some recent algorithmic advances in the domain.
We therefore use the \Int and \rocket  algorithms to cover two types of algorithmic families. 
It is important to note that these algorithms work in different ways.
Some, like \rocket, only play the role of feature extractor and must be coupled with a pure classifier, as ridge regression (RR)~\ref{tab:algo_type}.
%\ro{This choice of RR as the classifier to complement \texttt{rocket} is motivated by its robustness to high-dimensional data and its regularization capabilities.}
This choice of RR as the classifier to complement \texttt{rocket} is motivated by its robustness to high-dimensional data and its regularization capabilities.
On the other hand, other algorithms, such as \Int, play both roles directly.
Moreover, they are based on different techniques, as showed in Table~\ref{tab:method}.

\begin{table}[tb]
\caption{Task accomplished according to the algorithm used as baseline model for classification task}
\label{tab:algo_type}
\begin{center}
\scalebox{1.05}{% Adjust the scale value as needed
\begin{tabular}{|c||c|c|} 
\hline
\multicolumn{1}{|c||}{Algorithm} &  \multicolumn{1}{|c|}{Feature-Extractor} &  \multicolumn{1}{|c|}{Classifier}\\ 
\hline
 \rocket     &x&      \\  
\hline
\Int     &x&x    \\  
\hline
\end{tabular}
} % end scalebox
\end{center}
\end{table}

\begin{table}[tb]
\caption{Methodology based on the baseline classification algorithm employed. Since \texttt{ROCKET} functions primarily as a feature extractor, it is employed in conjunction with a Ridge Regressor (RR) for the classification task.}
\label{tab:method}
\begin{center}
\scalebox{1.00}{% Adjust the scale value as needed
\begin{tabular}{|c||c|c|c|} 
\hline
\multicolumn{1}{|c||}{Algorithm} &  \multicolumn{1}{|c|}{DL-based} & \multicolumn{1}{|c|}{Ensemble-based} &  \multicolumn{1}{|c|}{Kernel-based} \\ 
\hline
 \rocket  + RR   &  &   &x     \\  
\hline
\Int    &x&x&     \\  
\hline
\end{tabular}
} % end scalebox
\end{center}
\end{table}

%-----------------------------------------------------------------------------------------------------------
\subsection{Datasets and Experimental Settings}
\label{subsec:classification_experimental_settings}

\begin{table*}[tb]
\caption{Information about the original multivariate imbalanced datasets}
\label{info_data}
\begin{center}
{\scalebox{0.95}{% Adjust the scale value as needed
\begin{tabular}{lccccccccc}
\toprule
Dataset & n\_classes & Train\_size & Dim & Length & Var\_train & Var\_test & Im\_ratio & d\_train\_test & prop\_miss \\
\midrule
CharacterTrajectories          &        20 &       1422 &         3 &    182 &           0.15 &          0.15 &    13.06 &         3.35 &           0.33 \\
EigenWorms                     &         5 &           128 &         6 &  17984 &           0.18 &          0.18 &     3.26 &       386.95 &              0 \\
Epilepsy                       &         4 &           137 &         3 &    206 &           0.18 &          0.18 &     1.05 &         6.03 &              0 \\
EthanolConcentration           &         4 &           261 &         3 &   1751 &           0.24 &          0.23 &        2 &       101616 &              0 \\
FingerMovements                &         2 &           316 &        28 &     50 &           0.16 &          0.18 &        0 &       588.92 &              0 \\
Handwriting                    &        26 &           150 &         3 &    152 &           0.15 &           0.1 &    12.23 &         4.04 &              0 \\
Heartbeat                      &         2 &           204 &        61 &    405 &           0.09 &          0.09 &      0.3 &        23.15 &              0 \\
LSST                           &        14 &          2459 &         6 &     36 &           0.03 &          0.02 &     9.49 &      2259.42 &              0 \\
PEMS-SF                        &         7 &           267 &       963 &    144 &           0.17 &          0.18 &     3.07 &        30.79 &              0 \\
PenDigits                      &        10 &          7494 &         2 &      8 &            0.3 &          0.29 &     4.02 &        12.53 &              0 \\
RacketSports                   &         4 &           151 &         6 &     30 &           0.14 &          0.14 &     1.06 &        19.56 &              0 \\
SelfRegulationSCP1             &         2 &           268 &         6 &    896 &           0.16 &          0.15 &        0 &      3352.33 &              0 \\
SpokenArabicDigits             &        10 &          6599 &        13 &     93 &           0.14 &          0.13 &        0 &        38.48 &           0.57 \\

\bottomrule
\end{tabular}}
} % font size
\end{center}

\end{table*}

We evaluate the performance of baseline models and data augmentation techniques on the UCR/UEA archive~\cite{ucr,uea, bake_off_2017}, and use the 13 imbalanced multivariate datasets.

We use 5 different data augmentation techniques: a traditional noise injection with 3 different levels of noise $l \in \{1,3,5\}$, the SMOTE algorithm~\cite{SMOTE} and the TimeGANs~\cite{TimeGAN} generative algorithm. 
Injecting noise is known to be a reliable and fast augmentation technique, especially in computer vision.
Its use with 3 different levels also allows it to be used as a range method from the preserving technique branch.
SMOTE is a good representative of the interpolation-based techniques family, and TimeGANs are, to the best of our knowledge, the only generative model to take into account the temporal aspect of time series.
Note that among these three techniques, only the TimeGANs are time series-based and require external training. 
These techniques were used to augment the original or downsampled training set. 
The classification task is then performed by  \rocket  or \Int.

Table~\ref{tab:method} illustrates the diverse methodologies employed by the baseline algorithms in addressing the classification challenge. 
Among these, algorithms utilizing deep learning (DL) principles, particularly those based on residual neural networks, have found extensive application in tasks such as image recognition and classification~\cite{imagenet_classif, resnet_image}. 
The \rocket algorithm, notable for its innovative use of a vast number of randomly generated weights, aims to maximize the informational input to traditional classifiers, including logistic regression (LR) and ridge regression (RR).

To quantify the effectiveness of data augmentation, we introduce the concept of relative gain, $G_r$, defined as:
\begin{equation}
    G_r = \frac{\text{acc(model\_aug)} - \text{acc(model)}}{\text{acc(model)}},
\end{equation}
where \texttt{acc} represents the average accuracy over five runs, \texttt{model} denotes the model trained on the original dataset, and \texttt{model\_aug} signifies the same model trained on the augmented dataset.

Additionally, while some data characteristics are adopted from existing literature~\cite{survey_ts_classif}, we propose extensions and additions to these definitions to better accommodate the nuances of multivariate datasets. 
This expansion is critical for a comprehensive understanding of dataset attributes. 
For an exhaustive enumeration of these properties and their values across the 13 multivariate imbalanced datasets, refer to Table~\ref{info_data}.

\begin{itemize}
    \item Number of classes (n\_classes): The number of the classes present in the dataset. 
    \item Training set size (Train\_size): The number of time series in the original training set. 
    \item Dimension (Dim): The number of features in the dataset. 
    \item Time series length (Length): The length of time series in the original training set. 
    \item Dataset variance (Var\_train and Var\_test): To define a multivariate variance for our dataset, we consider the following equations:
\begin{align}
    \sigma_{mt}^2 &= \frac{1}{N} \sum_{i=1}^N (x_{imt} - \frac{1}{N} \sum_{i=1}^N x_{imt})^2, & \\
    \sigma_{D}^2 &= \frac{1}{T \times M} \sum_{m=1}^M \sum_{t=1}^{T} \sigma_{mt}^2 &
\end{align}

where $N$ is the number of time series in the dataset $D$, $M$ is the number of dimensions in each time series, $T$ is the length of the time series, and $x_{imt}$ denotes the value at time step $t$ of dimension $m$ in time series $i$. 
Furthermore, $\sigma_{mt}^2$ represents the variance at time step $t$ for dimension $m$ across the dataset, and $\sigma_{D}^2$ encapsulates the overall variance of the dataset $D$, averaged across all dimensions. 
    \item Class Imbalance (Im\_ratio): We used the imbalanced degree (ID) proposed by \cite{ortigosa} with Helliger distance, as recommended. 
    \item Train/Test distance(d\_train\_test): the Euclidean distance between the training set and the testing set. It is the Euclidean distance between the mean vector of the train and the test vector, the variance being already taken into account in another definition. This distance allows capturing a possible shift domain between the training set and the testing set. 
    \item Missing values proportion (prop\_miss) : Number of missing time steps divided by the total number of time steps in the dataset. 
\end{itemize} 

We did not consider the "\emph{patterns per class}" property, because~\cite{survey_ts_classif} showed that "\emph{the correlation of the change in accuracy to the average number of patterns per class is similar to training set size}" nor the intra-class variance, proportional to the variance of the dataset.

\subsection{Augmentation Protocol and Parameters}
\label{subsec:aug_protocol}

We have studied 13 multivariate datasets from the UCR/UEA archive, with different properties. 
The same division into training and testing sets was made as in the UCR/UEA archive. 
Among the imbalanced datasets, each one was previously augmented with one of the following techniques: timeGANs, SMOTE, noise\_1, noise\_3 and noise\_5 where $i$ in noise\_i refers to the standard deviation (std) multiplicator of noise, i.e. the level of injected noise $l \in \text{\{1,3,5\}}$. 
Indeed, we add to the dimension $j$ of the original time series a noise as the following:
\begin{equation}
    Noise \sim \mathcal{N}(0, l \times std_j)
\end{equation} where $std_j$ refers to the std of the dimension number $j$ of the original time series.
The addition of noise in a certain dimension is therefore proportional to the original std of this same dimension. 
For each class, we extract a time series randomly and add noise until the dataset is perfectly balanced.
For timeGANs, the number of iterations during training steps are set to 2500, 2500 and 1000 respectively. 
The dimension of the latent space is set to 10, gamma is set to 1, the learning rate to $5.10^{-4}$ and the batch size to 32.
We provide to the timeGANs, for each training,  time series coming from a single class of the original dataset, so that the generated series follow the same distribution, until the dataset is perfectly balanced. 
Concerning SMOTE, the number of neighbors to be considered is defined as the minimum between 5 and the number of elements in the class minus 1. 
The baseline models were applied on both the augmented and non-augmented datasets, i.e. 6 datasets per baseline model, and we compare the performance on each of them trying to capture some correlations between G and the aforementioned properties.

\begin{table*}[tb]
\caption{Accuracy for \texttt{rocket} baseline model, and relative improvement}
\label{rocket_results}
\begin{center}
\scalebox{0.97}{
\begin{tabular}{lccccccc}
\toprule
              Dataset &   \rocket & rocket\_noise\_1.0 &  rocket\_noise\_3.0 & rocket\_noise\_5.0 &   rocket\_smote & rocket\_timegan &  Improvement (\%) \\
\midrule
CharacterTrajectories &          98.52 &            99.09 &             99.04 &            99.12 &          98.47 & \textbf{99.19} &     0.68 \\
EigenWorms &          89.16 &            79.54 &             82.60 &            83.97 & \textbf{91.15} &          88.93 &     2.23 \\
Epilepsy &          98.99 &            98.12 &             98.41 &            98.26 &          98.55 & \textbf{99.28} &     0.29 \\
EthanolConcentration &          41.29 &            39.16 &             40.08 &            40.53 & \textbf{42.43} &          42.05 &     2.76 \\
FingerMovements &           52.20 &             54.80 &             54.00 &    \textbf{55.00} &           53.80 &           54.80 &     5.36 \\
Handwriting &          58.71 &            59.13 &             56.61 &            56.78 & \textbf{59.91} &          57.93 &     2.04 \\
Heartbeat &          73.76 &            73.07 &             74.63 &            72.59 & \textbf{75.32} &          74.34 &     2.11 \\
LSST & \textbf{63.84} &            61.97 &             62.54 &            62.64 &          61.39 &          63.78 &     -0.09 \\
PEMS-SF &          82.43 &   \textbf{83.93} &             82.66 &            83.35 &          83.35 &          82.31 &     1.82 \\
PenDigits & \textbf{97.87} &            97.77 &             97.75 &            97.71 &          97.72 &          97.66 &     -0.10 \\
RacketSports &          90.66 &            90.92 &             91.05 &            90.53 &          91.32 & \textbf{91.58} &     1.01 \\
SelfRegulationSCP1 & \textbf{85.39} &            84.85 &             85.19 &            85.19 &          84.51 &          84.98 &     -0.23 \\
SpokenArabicDigits &           96.20 &            98.34 &             98.23 &            98.26 &          96.44 &  \textbf{98.40} &     2.29 \\
\midrule
Average Improvement & - & - & - & - & -  &-  & 1.55 \\
\bottomrule
\end{tabular}
} % scalebox
\end{center}
\end{table*}

\begin{table*}[tb]

\caption{Accuracy for \Int (InT) baseline model, and relative improvement }
\label{fig:int_results}
\begin{center}
{\scalebox{0.97}{
\begin{tabular}{lccccccc}
\toprule
Dataset & \Int &  InT\_noise\_1.0 &  InT\_noise\_3.0 &  InT\_noise\_5.0 &  InT\_smote &  InT\_timegan & Improvement (\%) \\
\midrule
CharacterTrajectories          &   99.51 &          99.51 &          99.30 &          99.20 &      \textbf{99.55} &        99.41 & 0.04 \\
EigenWorms                     &   92.37 &          92.62 &          89.31 &          89.57 &      \textbf{94.66} &        86.77 & 2.48 \\
Epilepsy                       &   97.10 &          \textbf{97.39} &          96.81 &          96.96 &      97.25 &        96.96 & 0.30 \\
EthanolConcentration           &   23.19 &          24.33 &          20.15 &          22.81 &      \textbf{24.52} &        23.57 & 5.74 \\
FingerMovements                &   \textbf{53.20} &          50.40 &          48.60 &          47.80 &      51.00 &        48.40 & -4.14 \\
Handwriting                    &   \textbf{64.33} &          60.78 &          58.52 &          58.19 &      63.29 &        57.84 & -1.62 \\
Heartbeat                      &   71.22 &          71.41 &          \textbf{73.37} &          72.78 &      71.51 &        70.15 & 3.02 \\
LSST                           &   69.40 &          65.25 &          62.40 &          62.04 &      67.60 &        \textbf{69.91} & 0.73 \\
PEMS-SF                        &   \textbf{81.21} &          78.61 &          77.75 &          78.61 &      78.61 &        78.61 & -3.20 \\
PenDigits                      &   98.96 &          98.74 &          98.77 &          \textbf{98.99} &      \textbf{98.99} &        98.79 & 0.03 \\
RacketSports                   &   87.89 &          \textbf{89.80} &          \textbf{89.80} &          87.83 &      88.03 &        88.82 & 2.17 \\
SelfRegulationSCP1             &   76.18 &          74.74 &          76.25 &          76.25 &      \textbf{77.27} &        77.00 & 1.43 \\
SpokenArabicDigits             &   99.14 &          98.93 &          98.79 &          \textbf{99.41} &      98.93 &        98.98 & 0.27\\
\midrule
Average Improvement & - & - & - & - & -  &-  & 0.56 \\
\bottomrule
\end{tabular}
} % font size
}
\end{center}
\end{table*}

%-----------------------------------------------------------------------------------------------------------
\subsection{Classification Methodology and Setup}
\label{subsec:classif_protocol}
Our analysis employs two baseline models for evaluation: \Int and \rocket coupled with a ridge regression classifier. 
In the case of \rocket, we adhere to the default configuration, utilizing 10,000 kernels. 
For \Int, the dataset is partitioned into training and validation segments, maintaining a 2:1 ratio. Augmented data are incorporated exclusively during the training phase, ensuring the validation set comprises solely original, stratified samples. 
This approach aligns our evaluation with standard practices, facilitating direct comparison with other studies utilizing the complete UCR/UEA archive's test set.

Consistency in parameter settings is maintained across models, irrespective of augmentation, to ensure comparability. 
The training process extends over 200 epochs, incorporating an early stopping mechanism triggered after 30 epochs without improvement, preserving the best model based on validation accuracy. 
Prior to training, a cyclical learning rate analysis~\cite{cyclical_lr} is conducted for each dataset to identify the optimal learning rate, which is then adjusted to the identified valley point for subsequent training.

%-----------------------------------------------------------------------------------------------------------
\subsection{Effect of Augmentation on Model Classification}
\label{subsec:effects}

In this section, we show that data augmentation can effectively increase the accuracy performance of both classification models used. 
This is true, even in the cases where the original performance is already high.

The \rocket algorithm generates a large quantity of random convolutional kernels, all independent of each other~\cite{rocket}.
The latent space resulting from the extraction has a very large dimension some of which can be redundant, the information is therefore saturated. 

First, we note that on 10 out of the 13 datasets, the accuracy of the augmented models are better than those of the non-augmented model as shown in Table~\ref{rocket_results}.
This table shows an average relative improvement of 1.55\% across the 13 multivariate datasets when applying the best-performing data augmentation technique compared to the baseline \rocket classifier. 
It's observed that 6 out of the 13 datasets boast a baseline accuracy of 89\% or higher, making an average improvement of 1.55\%. 
Notably, in the 3 datasets where no improvement is seen, the augmented accuracies nearly match the baseline, representing the smallest absolute values among the 13 datasets, with a mere 0.14\% depreciation on these 3 datasets. 
Furthermore, among the 4 datasets with less than 2\% improvement, all have a baseline accuracy over 80\%, and out of the 6 datasets with a baseline accuracy of 89\% or more, 5 out of 6 show improvements. 
This highlights that data augmentation does not necessarily enhance the relative accuracies of datasets with already low performance with \rocket but also optimizes those with very high baseline accuracies. 
This observation underscores the complexity of time series classification tasks, and given the exceptionally high accuracy of the majority of datasets with a state-of-the-art model like \rocket, substantial improvements are not always expected.
It is also crucial to highlight that the effectiveness of augmentation techniques can vary across different datasets, suggesting that there is no one-size-fits-all solution for data augmentation in time series classification (see Table~\ref{tab:imp_occurences}).

\iffalse
\begin{figure}[tb]
  \includegraphics[width=\linewidth]{figures/rocket_diag.png}
  \caption{Accuracy of the best augmented \rocket model VS accuracy of the original \rocket model. \tp{increase size of all fonts; we cannot read anything!} \tp{I am not sure if this graph is supporting our argument: all points are on the diagonal, it seems as if there is no improvement; maybe maybe have a graph that reports the percent increase(/decrease) in accuracy?}} 
  \label{fig:rocket_diag}
\end{figure}
\fi

\iffalse
\begin{figure}[tb]
  \includegraphics[width=\linewidth]{figures/int_diag.png}
  \caption{Accuracy of the best augmented InceptionTime model VS accuracy of the original InceptionTime model. \tp{increase size of all fonts; we cannot read anything!} \tp{I am not sure if this graph is supporting our argument: all points are on the diagonal, it seems as if there is no improvement; maybe maybe have a graph that reports the percent increase(/decrease) in accuracy?}}
  \label{fig:int_diag}
\end{figure}
\fi

The \Int architecture, inspired by the success of Inception modules in image recognition, demonstrates significant efficacy in time series classification~\cite{inception_time}. 
Incorporating multiple Inception modules allows \Int to adeptly capture complex features from time series data at various scales. 

When examining the impact of data augmentation on \Int's performance, an average increase of 0.56\% in accuracy is observed across 10 out of the 13 multivariate datasets, as shown in Table~\ref{fig:int_results}. 
This improvement, though seemingly modest, underscores the potential of data augmentation to enhance the model's generalization from training data. 
Notably, all 7 datasets with a baseline accuracy of 87\% or higher experienced performance gains post-augmentation, highlighting a consistent benefit in scenarios where \Int already performs well—specifically, above the 85\% mark. 
This reveals an interesting pattern: data augmentation tends to yield advantages especially when the initial model performance is substantial. 
On the flip side, datasets with a relatively low baseline saw a negative impact from augmentation, suggesting that for deep learning models like InceptionTime, augmentation is more beneficial when starting performance is strong.

Moreover, Table~\ref{info_data} indicates that the 3 datasets without augmentation benefits have between 100 to 300 instances in total, with 30 to 150 time series per class. 
This points to an inherent data scarcity issue, particularly challenging for models that require extensive external training, such as TimeGANs. 
This pattern, akin to observations with the \rocket algorithm, highlights that while data augmentation can indeed refine a model's ability to generalize, its effectiveness varies across datasets. 
As depicted in Table~\ref{tab:imp_occurences}, diverse augmentation strategies contribute to performance improvements, emphasizing, as for \rocket, that there's no one-size-fits-all solution.

\subsection{Future Work}

Note that the contribution of data augmentation techniques to time series classification is fairly uniform. 
For instance, SMOTE contributes to improvements in 8 out of 13 cases for both \rocket and \Int models. TimeGANs shows effectiveness in 7 out of 13 cases for \rocket, and in 4 out of 13 for \Int. 
Noise augmentation presents improvements in 7 cases for \rocket and 8 for \Int. 
Note that simple techniques, like like SMOTE and Noise, show performance superior to TimeGAN in enhancing \Int (maybe due to the small training data sizes in our setting), suggesting that complex techniques are not always the most effective solution.

OVerall, the above results do not suggest a clear pattern that one could exploit to assert superiority of any specific augmentation technique over others. 
Furthermore, since a technique can perform well across datasets with different characteristics, it indicates the potential for combining techniques from various branches of our taxonomy. 
Similar to the augmentation pipelines in computer vision, where methods like CutMix~\cite{yun2019cutmix} are combined to enhance model performance, a conjunctive application of multiple time series augmentation methods could lead to further improvements.

\begin{table}[tb]
\caption{Count of Improvement Occurrences Over Baseline}
\label{tab:imp_occurences}
\begin{center}
\scalebox{1.0}{
\begin{tabular}{lcc}
\toprule
Augmentation Technique & \rocket & \Int \\
\midrule
SMOTE & 8 & 8 \\
TimeGAN & 7 & 4 \\
Noise & 7 & 8 \\
\bottomrule
\end{tabular}}
\label{tab:improvement_counts}
\end{center}
\end{table}

\section{Conclusions}
\label{sec:conclusion}
This study marks an advancement in the field of time series classification by incorporating a broad spectrum of data augmentation techniques, evaluated across both the \Int and \rocket models. 
By introducing a novel taxonomy of augmentation methods, we provide a structured approach to enhancing model performance in handling multivariate, imbalanced datasets. 
Our findings underscore the potential of data augmentation to improve accuracy, but demonstrate that no single technique consistently dominates across all datasets. 
This suggests that the strategic combination of diverse augmentation strategies, inspired by successful methodologies in computer vision, could lead to further improvements in model accuracy.
We hope our work paves the way for innovative approaches to leveraging these techniques for more robust and accurate models.

% Bibliography
\bibliographystyle{IEEEtranN}
\bibliography{refs}

% Generated by IEEEtranN.bst, version: 1.14 (2015/08/26)
\begin{thebibliography}{98}
\providecommand{\natexlab}[1]{#1}
\providecommand{\url}[1]{#1}
\csname url@samestyle\endcsname
\providecommand{\newblock}{\relax}
\providecommand{\bibinfo}[2]{#2}
\providecommand{\BIBentrySTDinterwordspacing}{\spaceskip=0pt\relax}
\providecommand{\BIBentryALTinterwordstretchfactor}{4}
\providecommand{\BIBentryALTinterwordspacing}{\spaceskip=\fontdimen2\font plus
\BIBentryALTinterwordstretchfactor\fontdimen3\font minus \fontdimen4\font\relax}
\providecommand{\BIBforeignlanguage}[2]{{%
\expandafter\ifx\csname l@#1\endcsname\relax
\typeout{** WARNING: IEEEtranN.bst: No hyphenation pattern has been}%
\typeout{** loaded for the language `#1'. Using the pattern for}%
\typeout{** the default language instead.}%
\else
\language=\csname l@#1\endcsname
\fi
#2}}
\providecommand{\BIBdecl}{\relax}
\BIBdecl

\bibitem[Olson et~al.(2018)Olson, Wyner, and Berk]{modern_nn}
M.~Olson, A.~Wyner, and R.~Berk, ``Modern neural networks generalize on small data sets,'' \emph{NeurIPS}, pp. 3619--3628, 2018.

\bibitem[Banko and Brill(2001)]{NLP}
M.~Banko and E.~Brill, ``Scaling to very very large corpora for natural language disambiguation,'' in \emph{Association for Computational Linguistic, 39th Annual Meeting and 10th Conference of the European Chapter, Proceedings of the Conference, July 9-11, 2001, Toulouse, France}.\hskip 1em plus 0.5em minus 0.4em\relax Morgan Kaufmann Publishers, 2001, pp. 26--33.

\bibitem[Shorten and Khoshgoftaar(2019)]{survey_image}
C.~Shorten and T.~M. Khoshgoftaar, ``A survey on image data augmentation for deep learning,'' \emph{J. Big Data}, vol.~6, no.~1, 2019.

\bibitem[Mikołajczyk and Grochowski(2018)]{dl_image_classif}
A.~Mikołajczyk and M.~Grochowski, ``Data augmentation for improving deep learning in image classification problem,'' \emph{IEEE}, 2018.

\bibitem[Feng et~al.(2021)Feng, Gangal, Wei, Chandar, Vosoughi, Mitamura, and Hovy]{data_aug_nlp}
S.~Y. Feng, V.~Gangal, J.~Wei, S.~Chandar, S.~Vosoughi, T.~Mitamura, and E.~H. Hovy, ``A survey of data augmentation approaches for {NLP},'' in \emph{Findings of the Association for Computational Linguistics: {ACL/IJCNLP} 2021, Online Event, August 1-6, 2021}, C.~Zong, F.~Xia, W.~Li, and R.~Navigli, Eds., vol. {ACL/IJCNLP} 2021, 2021, pp. 968--988.

\bibitem[Wen et~al.(2021)Wen, Sun, Yang, Song, Gao, Wang, and Xu]{ts_data_aug_dl_survey}
Q.~Wen, L.~Sun, F.~Yang, X.~Song, J.~Gao, X.~Wang, and H.~Xu, ``Time series data augmentation for deep learning: A survey,'' in \emph{Proceedings of the Thirtieth International Joint Conference on Artificial Intelligence}.\hskip 1em plus 0.5em minus 0.4em\relax International Joint Conferences on Artificial Intelligence Organization, aug 2021.

\bibitem[Iwana and Uchida(2021)]{survey_ts_classif}
B.~K. Iwana and S.~Uchida, ``An empirical survey of data augmentation for time series classification with neural networks,'' \emph{{PLOS} {ONE}}, vol.~16, no.~7, p. e0254841, 2021.

\bibitem[Lang et~al.(2021)Lang, Peng, Cui, Yang, and Guo]{fault_pred_aug}
P.~Lang, K.~Peng, J.~Cui, J.~Yang, and Y.~Guo, ``Data augmentation for fault prediction of aircraft engine with generative adversarial networks,'' in \emph{{CAA} Symposium on Fault Detection, Supervision, and Safety for Technical Processes, {SAFEPROCESS} 2021, Chengdu, China, December 17-18, 2021}, 2021, pp. 1--5.

\bibitem[Babaei et~al.(2019)Babaei, Chen, and Maul]{anomaly_aug}
K.~Babaei, Z.~Chen, and T.~Maul, ``Data augmentation by autoencoders for unsupervised anomaly detection,'' \emph{CoRR}, vol. abs/1912.13384, 2019.

\bibitem[Blagus and Lusa(2013)]{imbalanced_classes}
R.~Blagus and L.~Lusa, ``Smote for high-dimensional class-imbalanced data,'' \emph{BMC Bioinformatics}, vol.~14, no.~1, 2013.

\bibitem[Dau et~al.(2018)Dau, Bagnall, Kamgar, Yeh, Zhu, Gharghabi, Ratanamahatana, and Keogh]{ucr}
H.~A. Dau, A.~J. Bagnall, K.~Kamgar, C.~M. Yeh, Y.~Zhu, S.~Gharghabi, C.~A. Ratanamahatana, and E.~J. Keogh, ``The {UCR} time series archive,'' \emph{CoRR}, vol. abs/1810.07758, 2018.

\bibitem[Bagnall et~al.(2018)Bagnall, Dau, Lines, Flynn, Large, Bostrom, Southam, and Keogh]{uea}
A.~J. Bagnall, H.~A. Dau, J.~Lines, M.~Flynn, J.~Large, A.~Bostrom, P.~Southam, and E.~J. Keogh, ``The {UEA} multivariate time series classification archive, 2018,'' \emph{CoRR}, vol. abs/1811.00075, 2018.

\bibitem[Dempster et~al.(2020)Dempster, Petitjean, and Webb]{rocket}
A.~Dempster, F.~Petitjean, and G.~I. Webb, ``Rocket: exceptionally fast and accurate time series classification using random convolutional kernels,'' \emph{Data Min. Knowl. Discov.}, vol.~34, no.~5, pp. 1454--1495, 2020.

\bibitem[Pelletier and et~al.(2020)]{inception_time}
C.~Pelletier and D.~F.~S. et~al., ``Inceptiontime: Finding alexnet for time series classification,'' \emph{Data Min. Knowl. Discov.}, vol.~34, no.~6, pp. 1936--1962, 2020.

\bibitem[Bagnall et~al.(2017)Bagnall, Lines, Bostrom, Large, and Keogh]{bake_off_2017}
A.~J. Bagnall, J.~Lines, A.~Bostrom, J.~Large, and E.~J. Keogh, ``The great time series classification bake off: a review and experimental evaluation of recent algorithmic advances,'' \emph{. Data Min. Knowl. Discov.}, vol.~31, no.~3, pp. 606--660, 2017.

\bibitem[Ruiz and et~al.(2021)]{bake_off_2021}
A.~Ruiz and M.~F. et~al., ``The great multivariate time series classification bake-off: a review and experimental evaluation of recent algorithmic advances,'' \emph{Data Min. Knowl. Discov.}, vol.~35, pp. 401--449, 2021.

\bibitem[Shifaz et~al.(2019)Shifaz, Pelletier, Petitjean, and Webb]{ts_chief}
A.~Shifaz, C.~Pelletier, F.~Petitjean, and G.~I. Webb, ``{TS-CHIEF:} {A} scalable and accurate forest algorithm for time series classification,'' \emph{CoRR}, vol. abs/1906.10329, 2019.

\bibitem[Middlehurst et~al.(2021)Middlehurst, Large, Flynn, Lines, Bostrom, and Bagnall]{hive_cote_2}
M.~Middlehurst, J.~Large, M.~Flynn, J.~Lines, A.~Bostrom, and A.~J. Bagnall, ``{HIVE-COTE} 2.0: a new meta ensemble for time series classification,'' \emph{CoRR}, vol. abs/2104.07551, 2021.

\bibitem[Chawla et~al.(2002)Chawla, Bowyer, Hall, and Kegelmeyer]{SMOTE}
N.~V. Chawla, K.~W. Bowyer, L.~O. Hall, and W.~P. Kegelmeyer, ``{SMOTE}: Synthetic minority over-sampling technique,'' \emph{Journal of Artificial Intelligence Research}, vol.~16, pp. 321--357, 2002.

\bibitem[Yoon et~al.(2019)Yoon, Jarrett, and van~der Schaar]{TimeGAN}
J.~Yoon, D.~Jarrett, and M.~van~der Schaar, ``Time-series generative adversarial networks,'' in \emph{Advances in Neural Information Processing Systems}, H.~Wallach, H.~Larochelle, A.~Beygelzimer, F.~d\textquotesingle Alch\'{e}-Buc, E.~Fox, and R.~Garnett, Eds., vol.~32.\hskip 1em plus 0.5em minus 0.4em\relax Curran Associates, Inc., 2019.

\bibitem[Le~Guennec et~al.(2016)Le~Guennec, Malinowski, and Tavenard]{Data}
A.~Le~Guennec, S.~Malinowski, and R.~Tavenard, ``{Data Augmentation for Time Series Classification using Convolutional Neural Networks},'' in \emph{{ECML/PKDD Workshop on Advanced Analytics and Learning on Temporal Data}}, 2016.

\bibitem[Iwana and Uchida(2020)]{Guide}
B.~K. Iwana and S.~Uchida, ``Time series data augmentation for neural networks by time warping with a discriminative teacher,'' 2020.

\bibitem[Rashid and Louis(2019)]{time_warp}
K.~M. Rashid and J.~Louis, ``Time-warping: A time series data augmentation of imu data for construction equipment activity identification,'' in \emph{Proceedings of the 36th International Symposium on Automation and Robotics in Construction (ISARC)}, M.~Al-Hussein, Ed., 2019, pp. 651--657.

\bibitem[DeVries and Taylor(2017{\natexlab{a}})]{Cutout}
\BIBentryALTinterwordspacing
T.~DeVries and G.~W. Taylor, ``Improved regularization of convolutional neural networks with cutout,'' 2017. [Online]. Available: \url{https://arxiv.org/pdf/1708.04552v2.pdf}
\BIBentrySTDinterwordspacing

\bibitem[Park et~al.(2019)Park, Chan, and al.]{SpecAugment}
\BIBentryALTinterwordspacing
D.~S. Park, W.~Chan, and al., ``{SpecAugment}: A simple data augmentation method for automatic speech recognition,'' in \emph{Interspeech 2019}.\hskip 1em plus 0.5em minus 0.4em\relax {ISCA}, sep 2019. [Online]. Available: \url{https://doi.org/10.21437\%2Finterspeech.2019-2680}
\BIBentrySTDinterwordspacing

\bibitem[Matsuoka(1992)]{noise_injection}
K.~Matsuoka, ``Noise injection into inputs in back-propagation learning,'' \emph{IEEE Transactions on Systems, Man, and Cybernetics}, vol.~22, no.~3, pp. 436--440, 1992.

\bibitem[C.M.Bishop(1995)]{Tikhonov}
\BIBentryALTinterwordspacing
C.M.Bishop, ``Training with noise is equivalent to tikhonov regularization,'' \emph{Neural Computation}, 1995. [Online]. Available: \url{https://www.microsoft.com/en-us/research/wp-content/uploads/2016/02/bishop-tikhonov-nc-95.pdf}
\BIBentrySTDinterwordspacing

\bibitem[Greff et~al.(2017)Greff, Srivastava, Koutnik, Steunebrink, and Schmidhuber]{lstm}
\BIBentryALTinterwordspacing
K.~Greff, R.~K. Srivastava, J.~Koutnik, B.~R. Steunebrink, and J.~Schmidhuber, ``{LSTM}: A search space odyssey,'' \emph{{IEEE} Transactions on Neural Networks and Learning Systems}, vol.~28, no.~10, pp. 2222--2232, oct 2017. [Online]. Available: \url{https://doi.org/10.1109\%2Ftnnls.2016.2582924}
\BIBentrySTDinterwordspacing

\bibitem[Huang(2019)]{tdnn_lstm}
C.~Huang, ``Exploring effective data augmentation with {TDNN-LSTM} neural network embedding for speaker recognition,'' in \emph{{IEEE} Automatic Speech Recognition and Understanding Workshop, {ASRU} 2019, Singapore, December 14-18, 2019}.\hskip 1em plus 0.5em minus 0.4em\relax {IEEE}, 2019, pp. 291--295.

\bibitem[Steven~Eyobu and Han(2018)]{fourier_average}
O.~Steven~Eyobu and D.~S. Han, ``Feature representation and data augmentation for human activity classification based on wearable imu sensor data using a deep lstm neural network,'' \emph{Sensors}, vol.~18, no.~9, 2018.

\bibitem[Jaitly and Hinton(2013)]{VTLP}
N.~Jaitly and E.~Hinton, ``Vocal tract length perturbation (vtlp) improves speech recognition,'' 2013.

\bibitem[Takahashi et~al.(2016)Takahashi, Gygli, Pfister, and Van~Gool]{AER}
\BIBentryALTinterwordspacing
N.~Takahashi, M.~Gygli, B.~Pfister, and L.~Van~Gool, ``Deep convolutional neural networks and data augmentation for acoustic event detection,'' 2016. [Online]. Available: \url{https://arxiv.org/abs/1604.07160}
\BIBentrySTDinterwordspacing

\bibitem[Cui et~al.(2014)Cui, Goel, and Kingsbury]{SFM}
X.~Cui, V.~Goel, and B.~Kingsbury, ``Data augmentation for deep neural network acoustic modeling,'' \emph{2014 IEEE International Conference on Acoustics, Speech and Signal Processing (ICASSP)}, pp. 5582--5586, 2014.

\bibitem[Sinapiromsaran(2016)]{ANSMOT}
K.~Sinapiromsaran, ``Adaptive neighbor synthetic minority oversampling technique under 1nn outcast handling,'' 2016.

\bibitem[Tarawneh et~al.(2020)Tarawneh, Hassanat, Almohammadi, Chetverikov, and Bellinger]{SMOTEFUNA}
A.~S. Tarawneh, A.~B.~A. Hassanat, K.~Almohammadi, D.~Chetverikov, and C.~Bellinger, ``Smotefuna: Synthetic minority over-sampling technique based on furthest neighbour algorithm,'' \emph{IEEE Access}, vol.~8, pp. 59\,069--59\,082, 2020.

\bibitem[Han et~al.(2005)Han, Wang, and Mao]{borderline}
H.~Han, W.~Wang, and B.~Mao, ``Borderline-smote: {A} new over-sampling method in imbalanced data sets learning,'' in \emph{Advances in Intelligent Computing, International Conference on Intelligent Computing, {ICIC} 2005, Hefei, China, August 23-26, 2005, Proceedings, Part {I}}, ser. Lecture Notes in Computer Science, D.~Huang, X.~S. Zhang, and G.~Huang, Eds., vol. 3644.\hskip 1em plus 0.5em minus 0.4em\relax Springer, 2005, pp. 878--887.

\bibitem[He et~al.(2008)He, Bai, Garcia, and Li]{ADASYN}
H.~He, Y.~Bai, E.~A. Garcia, and S.~Li, ``Adasyn: Adaptive synthetic sampling approach for imbalanced learning,'' in \emph{2008 IEEE International Joint Conference on Neural Networks (IEEE World Congress on Computational Intelligence)}, 2008, pp. 1322--1328.

\bibitem[Bellinger et~al.(2019)Bellinger, Sharma, Japkowicz, and Zaiane]{SWIM}
C.~Bellinger, S.~Sharma, N.~Japkowicz, and O.~R. Zaiane, ``Framework for extreme imbalance classification: Swim—sampling with the majority class,'' \emph{Knowledge and Information Systems}, vol.~62, pp. 841--866, 2019.

\bibitem[Wen et~al.(2019)Wen, Gao, Song, Sun, Xu, and Zhu]{RobustSTL}
Q.~Wen, J.~Gao, X.~Song, L.~Sun, H.~Xu, and S.~Zhu, ``{RobustSTL}: {A} robust seasonal-trend decomposition algorithm for long time series,'' in \emph{The Thirty-Third {AAAI} Conference on Artificial Intelligence, {AAAI} 2019, The Thirty-First Innovative Applications of Artificial Intelligence Conference, {IAAI} 2019, The Ninth {AAAI} Symposium on Educational Advances in Artificial Intelligence, {EAAI} 2019, Honolulu, Hawaii, USA, January 27 - February 1, 2019}, 2019, pp. 5409--5416.

\bibitem[Huang et~al.(1998)Huang, Shen, Long, Wu, Shih, Zheng, Yen, Tung, and Liu]{EMD}
N.~E. Huang, Z.~Shen, S.~R. Long, M.~C. Wu, H.~H. Shih, Q.~Zheng, N.~Yen, C.~C. Tung, and H.~H. Liu, ``The empirical mode decomposition and the hilbert spectrum for nonlinear and non-stationary time series analysis,'' \emph{Proceedings of the Royal Society of London. Series A: Mathematical, Physical and Engineering Sciences}, vol. 454, pp. 903 -- 995, 1998.

\bibitem[Nam et~al.(2020)Nam, Bu, Park, Seo, Jo, and Jeong]{emd_aug}
G.-H. Nam, S.-J. Bu, N.-M. Park, J.-Y. Seo, H.-C. Jo, and W.-T. Jeong, ``Data augmentation using empirical mode decomposition on neural networks to classify impact noise in vehicle,'' \emph{IEEE ICASSP}, 2020.

\bibitem[Gao et~al.(2020)Gao, Song, Wen, Wang, Sun, and Xu]{Robusttad}
\BIBentryALTinterwordspacing
J.~Gao, X.~Song, Q.~Wen, P.~Wang, L.~Sun, and H.~Xu, ``Robusttad: Robust time series anomaly detection via decomposition and convolutional neural networks,'' 2020. [Online]. Available: \url{https://arxiv.org/abs/2002.09545}
\BIBentrySTDinterwordspacing

\bibitem[Eltoft(2002)]{ICA_aug}
T.~Eltoft, ``Data augmentation using a combination of independent component analysis and non-linear time-series prediction,'' \emph{IJCNN}, p. 448–453, 2002.

\bibitem[Comon(1994)]{ICA}
P.~Comon, ``Independent component analysis, a new concept?'' \emph{Sig. Process}, vol.~36, no.~3, pp. 287--314, 1994.

\bibitem[Tanner and Wong(1987)]{posterior}
M.~A. Tanner and W.~H. Wong, ``The calculation of posterior distributions by data augmentation,'' \emph{J. American Stat. Assoc}, vol.~82, no. 398, pp. 528--540, 1987.

\bibitem[Cao et~al.(2014)Cao, Tan, and Pang]{Mixture}
H.~Cao, V.~Y. Tan, and J.~Z. Pang, ``A parsimonious mixture of gaussian trees model for oversampling in imbalanced and multimodal time-series classification,'' \emph{IEEE TNNLS}, p. 2226–2239, 2014.

\bibitem[Smyl and Kuber(2016)]{LGT}
S.~Smyl and K.~Kuber, ``Data preprocessing and augmentation for multiple short time series forecasting with recurrent neural networks,'' \emph{ISF}, 2016.

\bibitem[Kang et~al.(2020)Kang, Hyndman, and Li]{GRATIS}
\BIBentryALTinterwordspacing
Y.~Kang, R.~J. Hyndman, and F.~Li, ``{GRATIS}: {GeneRAting} {TIme} series with diverse and controllable characteristics,'' \emph{Statistical Analysis and Data Mining: The {ASA} Data Science Journal}, vol.~13, no.~4, pp. 354--376, may 2020. [Online]. Available: \url{https://doi.org/10.1002\%2Fsam.11461}
\BIBentrySTDinterwordspacing

\bibitem[Bengio et~al.(2012)Bengio, Mesnil, Dauphin, and Rifai]{latent_efficient}
\BIBentryALTinterwordspacing
Y.~Bengio, G.~Mesnil, Y.~Dauphin, and S.~Rifai, ``Better mixing via deep representations,'' \emph{ICML}, pp. 552--560, 2012. [Online]. Available: \url{https://arxiv.org/pdf/1207.4404.pdf}
\BIBentrySTDinterwordspacing

\bibitem[DeVries and Taylor(2017{\natexlab{b}})]{latent_space}
\BIBentryALTinterwordspacing
T.~DeVries and G.~W. Taylor, ``Dataset augmentation in feature space,'' \emph{ICLR 2017}, pp. 1--12, 2017. [Online]. Available: \url{https://arxiv.org/abs/1702.05538}
\BIBentrySTDinterwordspacing

\bibitem[Verma et~al.(2019)Verma, Lamb, Beckham, Najafi, Mitliagkas, Lopez-Paz, and Bengio]{mixup}
\BIBentryALTinterwordspacing
V.~Verma, A.~Lamb, C.~Beckham, A.~Najafi, I.~Mitliagkas, D.~Lopez-Paz, and Y.~Bengio, ``Manifold mixup: Better representations by interpolating hidden states,'' \emph{Proceedings of Machine Learning Research}, vol.~97, pp. 6438--6447, 2019. [Online]. Available: \url{https://arxiv.org/abs/1806.05236}
\BIBentrySTDinterwordspacing

\bibitem[Cheung and Yeung(2021)]{modals}
T.-H. Cheung and D.-Y. Yeung, ``Modality-agnostic automated data augmentation in the latent space,'' \emph{International Conference on Learning Representations (ICLR)}, 2021.

\bibitem[Tu et~al.(2018)Tu, Liu, Meng, Liu, and Ding]{LSTMAE}
J.~Tu, H.~Liu, F.~Meng, M.~Liu, and R.~Ding, ``Spatial-temporal data augmentation based on lstm autoencoder network for skeleton-based human action recognition,'' in \emph{2018 25th IEEE International Conference on Image Processing (ICIP)}, 2018, pp. 3478--3482.

\bibitem[Yang et~al.(2021)Yang, Zhang, Zhang, Zhao, and Cui]{DTWSSE}
\BIBentryALTinterwordspacing
X.~Yang, X.~Zhang, Z.~Zhang, Y.~Zhao, and R.~Cui, ``{DTWSSE}: Data augmentation with a~siamese encoder for time series,'' in \emph{Web and Big Data}.\hskip 1em plus 0.5em minus 0.4em\relax Springer International Publishing, 2021, pp. 435--449. [Online]. Available: \url{https://doi.org/10.1007\%2F978-3-030-85896-4_34}
\BIBentrySTDinterwordspacing

\bibitem[Fu et~al.(2020)Fu, Kirchbuchner, and Kuijper]{data_aug_ts_gen}
B.~Fu, F.~Kirchbuchner, and A.~Kuijper, ``Data augmentation for time series: traditional vs generative models on capacitive proximity time series,'' \emph{Proceedings of the 13th ACM International Conference on PErvasive Technologies Related to Assistive Environments}, 2020.

\bibitem[Goodfellow et~al.(2014)Goodfellow, Pouget-Abadie, Mirza, Xu, Warde-Farley, Ozair, Courville, and Bengio]{GAN}
I.~J. Goodfellow, J.~Pouget-Abadie, M.~Mirza, B.~Xu, D.~Warde-Farley, S.~Ozair, A.~Courville, and Y.~Bengio, ``Generative adversarial networks,'' \emph{CoRR}, vol. abs/1406.2661, 2014.

\bibitem[Madhu and Kumaraswamy(2019)]{WaveGAN}
A.~Madhu and S.~K. Kumaraswamy, ``Data augmentation using generative adversarial network for environmental sound classification,'' \emph{2019 27th European Signal Processing Conference (EUSIPCO)}, pp. 1--5, 2019.

\bibitem[Arjovsky et~al.(2017)Arjovsky, Chintala, and Bottou]{wgan_first}
M.~Arjovsky, S.~Chintala, and L.~Bottou, ``Wasserstein {GAN},'' \emph{CoRR}, vol. abs/1701.07875, 2017.

\bibitem[Luo et~al.(2018)Luo, Zhu, Wan, and Lu]{WGAN-VAEs}
Y.~Luo, L.-Z. Zhu, Z.-Y. Wan, and B.-L. Lu, ``Data augmentation for eeg-based emotion recognition with deep convolutional neural networks,'' \emph{ICMM}, pp. 82--93, 2018.

\bibitem[Ramponi et~al.(2018)Ramponi, Protopapas, Brambilla, and Janssen]{TCGAN}
G.~Ramponi, P.~Protopapas, M.~Brambilla, and R.~Janssen, ``{T-CGAN:} conditional generative adversarial network for data augmentation in noisy time series with irregular sampling,'' \emph{CoRR}, vol. abs/1811.08295, 2018.

\bibitem[Harada et~al.(2018)Harada, Hayashi, and Uchida]{bio_GANS}
S.~Harada, H.~Hayashi, and S.~Uchida, ``Biosignal data augmentation based on generative adversarial networks,'' \emph{2018 40th Annual International Conference of the IEEE Engineering in Medicine and Biology Society (EMBC)}, pp. 368--371, 2018.

\bibitem[Chen et~al.(2019)Chen, Zhu, Hong, and Yang]{emotionalGAN}
G.~Chen, Y.~Zhu, Z.~Hong, and Z.~Yang, ``Emotionalgan: Generating ecg to enhance emotion state classification,'' \emph{Proceedings of the 2019 International Conference on Artificial Intelligence and Computer Science}, 2019.

\bibitem[Oord et~al.(2016)Oord, Dieleman, Zen, Simonyan, Vinyals, Graves, Kalchbrenner, Senior, and Kavukcuoglu]{Wavenet}
A.~v.~d. Oord, S.~Dieleman, H.~Zen, K.~Simonyan, O.~Vinyals, A.~Graves, N.~Kalchbrenner, A.~Senior, and K.~Kavukcuoglu, ``Wavenet: A generative model for raw audio,'' 2016.

\bibitem[Salinas et~al.(2017)Salinas, Flunkert, and Gasthaus]{deep_ar}
D.~Salinas, V.~Flunkert, and J.~Gasthaus, ``Deepar: Probabilistic forecasting with autoregressive recurrent networks,'' 2017.

\bibitem[Benton et~al.(2022)Benton, Shi, Bortoli, Deligiannidis, and Doucet]{diffusion}
J.~Benton, Y.~Shi, V.~D. Bortoli, G.~Deligiannidis, and A.~Doucet, ``From denoising diffusions to denoising markov models,'' \emph{CoRR}, vol. abs/2211.03595, 2022.

\bibitem[Kobyzev et~al.(2021)Kobyzev, Prince, and Brubaker]{normalizing_flows}
I.~Kobyzev, S.~J. Prince, and M.~A. Brubaker, ``Normalizing flows: An introduction and review of current methods,'' \emph{{IEEE} Transactions on Pattern Analysis and Machine Intelligence}, vol.~43, no.~11, pp. 3964--3979, nov 2021.

\bibitem[Deng et~al.(2020)Deng, Chang, Brubaker, Mori, and Lehrmann]{dynamic_norm_flows}
R.~Deng, B.~Chang, M.~A. Brubaker, G.~Mori, and A.~Lehrmann, ``Modeling continuous stochastic processes with dynamic normalizing flows,'' \emph{NeurIPS 2020}, 2020.

\bibitem[Morrow and Chiu(2020)]{VAEs_flow}
R.~Morrow and W.~Chiu, ``Variational autoencoders with normalizing flow decoders,'' \emph{CoRR}, vol. abs/2004.05617, 2020.

\bibitem[Holmstrom and Koistinen(1992)]{noise_standard}
\BIBentryALTinterwordspacing
L.~Holmstrom and P.~Koistinen, ``Using additive noise in back-propagation training,'' \emph{Trans. Neur. Netw.}, vol.~3, no.~1, p. 24–38, jan 1992. [Online]. Available: \url{https://doi.org/10.1109/72.105415}
\BIBentrySTDinterwordspacing

\bibitem[Um et~al.(2017)Um, Pfister, and al.]{parkinson}
T.~T. Um, Pfister, and al., ``Data augmentation of wearable sensor data for parkinson’s disease monitoring using convolutional neural networks,'' in \emph{Proceedings of the 19th ACM International Conference on Multimodal Interaction}, 2017, p. 216–220.

\bibitem[Kim and Jeong(2021)]{class_preserv}
\BIBentryALTinterwordspacing
M.~Kim and C.~Y. Jeong, ``Label-preserving data augmentation for mobile sensor data,'' \emph{Multidimensional Syst. Signal Process.}, vol.~32, no.~1, p. 115–129, jan 2021. [Online]. Available: \url{https://doi.org/10.1007/s11045-020-00731-2}
\BIBentrySTDinterwordspacing

\bibitem[H.~Cao and Ng(2011)]{SPO}
Y.-K.~W. H.~Cao, X.-L.~Li and S.-K. Ng, ``Spo: Structure preserving oversampling for imbalanced time series classification,'' \emph{ICDM}, p. 1008–1013, 2011.

\bibitem[Cao and Ng(2013)]{INOS}
D.~Y.-K.~W. Cao, X.-L.~Li and S.-K. Ng, ``Integrated oversampling for imbalanced time series classification,'' \emph{IEEE Transactions on Knowledge and Data Engineering}, vol.~25, no.~12, p. 2809–2822, 2013.

\bibitem[Abdi and Hashemi(2016)]{MDO}
L.~Abdi and S.~Hashemi, ``To combat multi-class imbalanced problems by means of over-sampling techniques,'' \emph{IEEE Transactions on Knowledge and Data Engineering}, vol.~28, no.~1, pp. 238--251, 2016.

\bibitem[Zhu et~al.(2020)Zhu, Lin, and Liu]{oversampling_imbalanced}
T.~Zhu, Y.~Lin, and Y.~Liu, ``Oversampling for imbalanced time series data,'' \emph{CoRR}, vol. abs/2004.06373, 2020.

\bibitem[Sakoe and Chiba(1978)]{DTW}
H.~Sakoe and S.~Chiba, ``Dynamic programming algorithm optimization for spoken word recognition,'' \emph{IEEE Transactions on Acoustics, Speech, and Signal Processing}, vol.~26, pp. 159--165, 1978.

\bibitem[Zhao and Itti(2018)]{ShapeDTW}
J.~Zhao and L.~Itti, ``{shapeDTW}: Shape dynamic time warping,'' \emph{Pattern Recognit.}, vol.~74, pp. 171--184, 2018.

\bibitem[Petitjean et~al.(2011)Petitjean, Ketterlin, and Gançarski]{dba}
F.~Petitjean, A.~Ketterlin, and P.~Gançarski, ``A global averaging method for dynamic time warping, with applications to clustering,'' \emph{Pattern Recogn.}, vol.~44, no.~3, p. 678–693, 2011.

\bibitem[Bellman(1957)]{Bellman}
Bellman, \emph{Dynamic programming}, 1957.

\bibitem[Vinod et~al.(2009)Vinod, D, and Javier Lopez-de Lacalle]{Bootstrap}
Vinod, H.~D, and e.~a. Javier Lopez-de Lacalle, ``Maximum entropy bootstrap for time series: the meboot r package,'' \emph{Journal of Statistical Software}, vol.~29, no.~5, pp. 1--19, 2009.

\bibitem[Javeri et~al.(2021)Javeri, Toutiaee, Arpinar, Miller, and Miller]{NNAutoML}
I.~Y. Javeri, M.~Toutiaee, I.~B. Arpinar, T.~W. Miller, and J.~A. Miller, ``Improving neural networks for time series forecasting using data augmentation and automl,'' 2021.

\bibitem[Hasibi et~al.(2019)Hasibi, Shokri, and Fooladi]{traffic}
R.~Hasibi, M.~Shokri, and M.~D.~T. Fooladi, ``Augmentation scheme for dealing with imbalanced network traffic classification using deep learning,'' \emph{CoRR}, vol. abs/1901.00204, 2019.

\bibitem[Lou et~al.(2018)Lou, Qi, and Li]{WGAN}
H.~Lou, Z.~Qi, and J.~Li, ``One-dimensional data augmentation using a wasserstein generative adversarial network with supervised signal,'' in \emph{2018 Chinese Control And Decision Conference (CCDC)}, 2018, pp. 1896--1901.

\bibitem[Lim et~al.(2018)Lim, Loo, Tran, Cheung, Roig, and Elovici]{doping}
\BIBentryALTinterwordspacing
S.~K. Lim, Y.~Loo, N.~Tran, N.~Cheung, G.~Roig, and Y.~Elovici, ``{DOPING:} generative data augmentation for unsupervised anomaly detection with {GAN},'' \emph{CoRR}, vol. abs/1808.07632, 2018. [Online]. Available: \url{http://arxiv.org/abs/1808.07632}
\BIBentrySTDinterwordspacing

\bibitem[Makhzani et~al.(2015)Makhzani, Shlens, Jaitly, and Goodfellow]{AAE}
\BIBentryALTinterwordspacing
A.~Makhzani, J.~Shlens, N.~Jaitly, and I.~J. Goodfellow, ``Adversarial autoencoders,'' \emph{CoRR}, vol. abs/1511.05644, 2015. [Online]. Available: \url{http://arxiv.org/abs/1511.05644}
\BIBentrySTDinterwordspacing

\bibitem[Alexandrov et~al.(2019)Alexandrov, Benidis, Bohlke{-}Schneider, Flunkert, Gasthaus, Januschowski, Maddix, Rangapuram, Salinas, Schulz, Stella, T{\"{u}}rkmen, and Wang]{ts_modeling}
A.~Alexandrov, K.~Benidis, M.~Bohlke{-}Schneider, V.~Flunkert, J.~Gasthaus, T.~Januschowski, D.~C. Maddix, S.~S. Rangapuram, D.~Salinas, J.~Schulz, L.~Stella, A.~C. T{\"{u}}rkmen, and Y.~Wang, ``Gluonts: Probabilistic time series models in python,'' \emph{CoRR}, vol. abs/1906.05264, 2019.

\bibitem[Ye and Keogh(2009)]{shapelet}
L.~Ye and E.~J. Keogh, ``Time series shapelets: a new primitive for data mining,'' in \emph{Proceedings of the 15th {ACM} {SIGKDD} International Conference on Knowledge Discovery and Data Mining, Paris, France, June 28 - July 1, 2009}, J.~F.~E. IV, F.~Fogelman{-}Souli{\'{e}}, P.~A. Flach, and M.~J. Zaki, Eds., 2009, pp. 947--956.

\bibitem[Jarvis and Patrick(1973)]{SNN}
R.~A. Jarvis and E.~A. Patrick, ``Clustering using a similarity measure based on shared near neighbors,'' \emph{IEEE Transactions on Computers}, vol. C-22, pp. 1025--1034, 1973.

\bibitem[Fawaz and et~al.(2019)]{dl_classif_review}
H.~I. Fawaz and G.~F. et~al., ``Deep learning for time series classification: a review,'' \emph{Data Min. Knowl. Discov.}, vol.~33, no.~4, pp. 917--963, 2019.

\bibitem[Bagnall and et~al.(2015)]{cote}
A.~Bagnall and J.~L. et~al., ``Time-series classification with cote: the collective of transformation-based ensembles,'' \emph{IEEE Trans Knowl Data Eng}, vol.~27, p. 2522–2535, 2015.

\bibitem[Wang~Z(2017b)]{resnet}
O.~T. Wang~Z, Yan~W, ``Time series classification from scratch with deep neural networks: A strong baseline,'' \emph{International Joint Conference on Neural Networks}, p. 1578–1585, 2017b.

\bibitem[Lines~J(2016)]{hive_cote_2016}
B.~A. Lines~J, Taylor~S, ``Hive-cote: The hierarchical vote collective of transformation-based ensembles for time series classification,'' \emph{IEEE International Conference on Data Mining}, p. 1041–1046, 2016.

\bibitem[Lines~J(2018)]{hive_cote_2018}
------, ``Time series classification with hive-cote: The hierarchical vote collective of transformation-based ensembles,'' \emph{ACM Transactions on Knowledge Discovery from Data}, vol.~12, no.~5, 2018.

\bibitem[A.~Krizhevsky and Hinton(2012)]{imagenet_classif}
I.~S. A.~Krizhevsky and G.~E. Hinton, ``Imagenet classification with deep convolutional neural networks,'' \emph{NiPS}, p. 1097–1105, 2012.

\bibitem[He et~al.(2016)He, Zhang, Ren, and Sun]{resnet_image}
K.~He, X.~Zhang, S.~Ren, and J.~Sun, ``Deep residual learning for image recognition,'' in \emph{2016 {IEEE} Conference on Computer Vision and Pattern Recognition, {CVPR} 2016, Las Vegas, NV, USA, June 27-30, 2016}, 2016, pp. 770--778.

\bibitem[Ortigosa{-}Hern{\'{a}}ndez et~al.(2017)Ortigosa{-}Hern{\'{a}}ndez, Inza, and Lozano]{ortigosa}
J.~Ortigosa{-}Hern{\'{a}}ndez, I.~Inza, and J.~A. Lozano, ``Measuring the class-imbalance extent of multi-class problems,'' \emph{Pattern Recognit. Lett.}, vol.~98, pp. 32--38, 2017.

\bibitem[Smith(2017)]{cyclical_lr}
L.~N. Smith, ``Cyclical learning rates for training neural networks,'' in \emph{2017 {IEEE} Winter Conference on Applications of Computer Vision, {WACV} 2017, Santa Rosa, CA, USA, March 24-31, 2017}, 2017, pp. 464--472.

\bibitem[Yun et~al.(2019)Yun, Han, Oh, Chun, Choe, and Yoo]{yun2019cutmix}
S.~Yun, D.~Han, S.~J. Oh, S.~Chun, J.~Choe, and Y.~Yoo, ``Cutmix: Regularization strategy to train strong classifiers with localizable features,'' in \emph{Proceedings of the IEEE/CVF International Conference on Computer Vision (ICCV)}, 2019, pp. 6023--6032.

\end{thebibliography}
\end{document}